\pdfoutput=1

\documentclass[11pt]{article}

\usepackage[preprint]{acl}

\usepackage{times}
\usepackage{latexsym}
\usepackage{colortbl}
\usepackage[T1]{fontenc}

\usepackage[utf8]{inputenc}

\usepackage{microtype}

\usepackage{inconsolata}

\usepackage{graphicx}

\usepackage{amsmath,amsthm,amssymb,amsfonts}
\usepackage{algorithm}
\usepackage{algorithmic}
\usepackage{multirow}
\usepackage{enumitem}
\usepackage{tabularx}
\usepackage{pifont}
\usepackage{bbm}
\usepackage[most]{tcolorbox} 
\usepackage{lipsum}
\usepackage{booktabs} 
\title{\textsc{TestAgent}: Automatic Benchmarking and Exploratory Interaction for Evaluating LLMs in Vertical Domains}

\author{Wanying Wang$^{1,*}$, Zeyu Ma$^{2,1,*}$, Xuhong Wang$^{3}$,  Yangchun Zhang$^{4}$,\\ \textbf{Pengfei Liu$^{5,\dagger}$, Mingang Chen$^{1,\dagger}$} \\
$^{1}$ Shanghai Key Laboratory of Computer Software Testing and Evaluating,\\ Shanghai Development Center of Computer Software Technology\\
$^{2}$ Shanghai Normal University~~~~
$^{3}$ Shanghai Artificial Intelligence Lab\\
$^{4}$ Shanghai University~~~~
$^{5}$ Shanghai Jiao Tong University
}

\begin{document}
\maketitle
\begin{abstract}
As Large Language Models (LLMs) are increasingly deployed in highly specialized vertical domains, the evaluation of their domain-specific performance becomes critical. However, existing evaluations for vertical domains typically rely on the labor-intensive construction of static single-turn datasets, which present two key limitations: (i) manual data construction is costly and must be repeated for each new domain, and (ii) static single-turn evaluations are misaligned with the dynamic multi-turn interactions in real-world applications, limiting the assessment of professionalism and stability. To address these, we propose \textbf{\textsc{TestAgent}}, a framework for automatic benchmarking and exploratory dynamic evaluation in vertical domains. \textsc{TestAgent} leverages retrieval-augmented generation to create domain-specific questions from user-provided knowledge sources, combined with a two-stage criteria generation process, thereby enabling scalable and automated benchmark creation. Furthermore, it introduces a reinforcement learning–guided multi-turn interaction strategy that adaptively determines question types based on real-time model responses, dynamically probing knowledge boundaries and stability. Extensive experiments across medical, legal, and governmental domains demonstrate that \textsc{TestAgent} enables efficient cross-domain benchmark generation and yields deeper insights into model behavior through dynamic exploratory evaluation. This work establishes a new paradigm for automated and in-depth evaluation of LLMs in vertical domains.
\end{abstract}

\section{Introduction}
Large language models (LLMs) \cite{bubeck2023sparksartificialgeneralintelligence,llama,qwen3,gpt4o} have demonstrated remarkable capabilities across a wide range of tasks, accelerating the development of various domain applications \cite{bommasani2021opportunitiess,medical_survey}. As LLMs are increasingly deployed in highly specialized vertical domains, such as medicine \cite{medical_survey}, law \cite{yao2022leven}, and finance \cite{wu2023bloomberggpt}, evaluation demands have shifted beyond general-purpose capabilities, highlighting the need for rigorous assessments of domain-specific performance \cite{peng2024survey,guo_survey}.

Despite the progress in LLM evaluation, benchmarking in vertical domains remains challenging. Existing benchmarks primarily target general domains such as mathematical reasoning \cite{math,GSM8K} and common-sense knowledge \cite{mmlu,cmmlu}, with limited benchmarks tailored to professional, vertical domains. Some efforts have collected domain-specific datasets, such as LegalBench for legal domains \cite{guha2023legalbench} and MedBench for healthcare \cite{liu2024medbench}. However, these benchmarks rely heavily on manual data collection, which is labor-intensive and struggles to scale when new knowledge sources emerge. Consequently, when facing new domains, the entire benchmark construction process must restart from scratch, rendering efficient assessment infeasible.

Furthermore, a critical limitation of existing vertical benchmarks is their single-turn, static formats. Such formats are fundamentally misaligned with the nature of vertical domains for two reasons. First, knowledge in these domains is often deeply interrelated, necessitating multi-turn interactions that systematically validate the depth of model expertise. Second, stability is important in vertical scenarios, particularly given their frequent use in professional consultations, where contradictory responses can severely impair user trust and model credibility. These underscore the importance of evaluating the professionalism and stability of the model within dynamic, multi-turn interactions. However, manual construction of such interactive dialogues would greatly exacerbate evaluation costs, while dynamic methods like Dyval \cite{dyval} for general reasoning tasks, or adaptation-based approaches such as KIEval \cite{kieval} and VLB \cite{vlb}, are not equipped to automatically construct benchmarks from scratch for vertical domains. These challenges collectively highlight the need for an automated and scalable solution for dynamic multi-turn benchmarking tailored to vertical domains.
\begin{figure}[t]
    \centering
    \includegraphics[width=0.75\linewidth]{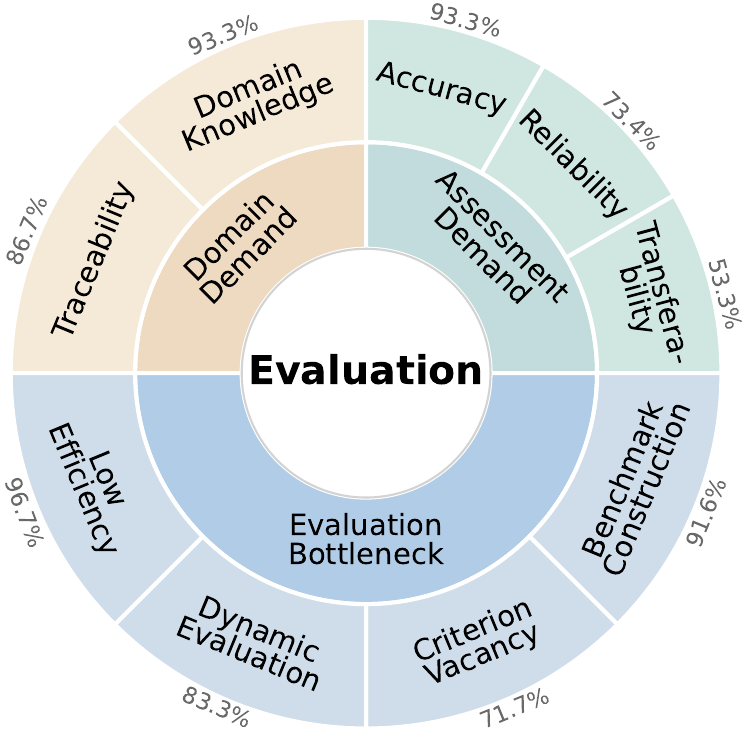}
    \caption{Statistical results of evaluation demand and bottleneck in the LLMs industry.}
    \label{fig:survey}
\end{figure}

Industry practices further confirm the urgency of vertical domain evaluation. As shown in Figure \ref{fig:survey}, our survey of 50 LLM companies (detailed in Appendix \ref{appendix:evaluation subjects}) reveals that 93.3\% prioritize domain-specific evaluation. However, over 90\% of evaluation providers struggle with constructing efficient domain-specific benchmarks, and more than 70\% report the need for dynamic evaluation with tailored criteria. This gap between evaluation demand and support significantly hinders the development and deployment of LLMs in vertical domains.

In this paper, we propose \textsc{TestAgent}, an agent-based framework designed for automated benchmarking and exploratory evaluation in vertical domains. Specifically, leveraging Retrieval-Augmented Generation (RAG), it first generates questions from user-provided knowledge sources in the given vertical domain. Subsequently, it employs a two-stage criteria generation method, initially generating coarse topic-level criteria via in-context learning and then refining them through retrieving question-relevant information. Building upon this, \textsc{TestAgent} introduces a novel reinforcement-learning (RL)-guided multi-turn dynamic interaction approach, intelligently selecting question types based on the target model's responses. This multi-turn evaluation facilitates in-depth analysis of domain expertise and stability, delivering comprehensive metrics aligned closely with real-world requirements.

Our contributions are summarized as follows:
\begin{itemize}[left=0pt]
\item We propose an automatic benchmark generation method for vertical domains based on RAG, enabling cross-domain adaptability and significantly enhancing benchmarking efficiency.
\item We introduce a novel RL-driven, multi-turn evaluation mechanism that adaptively generates contextually relevant questions based on model feedback, supporting in-depth assessment of domain knowledge and stability.
\item We integrate these components into \textsc{TestAgent}, an agent-based evaluation framework that supports automatic, cross-domain benchmark generation and dynamic evaluation. Extensive experiments in the medical, legal, and government domains demonstrate the adaptability and discriminative power of \textsc{TestAgent}.
\end{itemize}
    
\begin{figure*}[t]
\centering 
  \includegraphics[width=\linewidth]{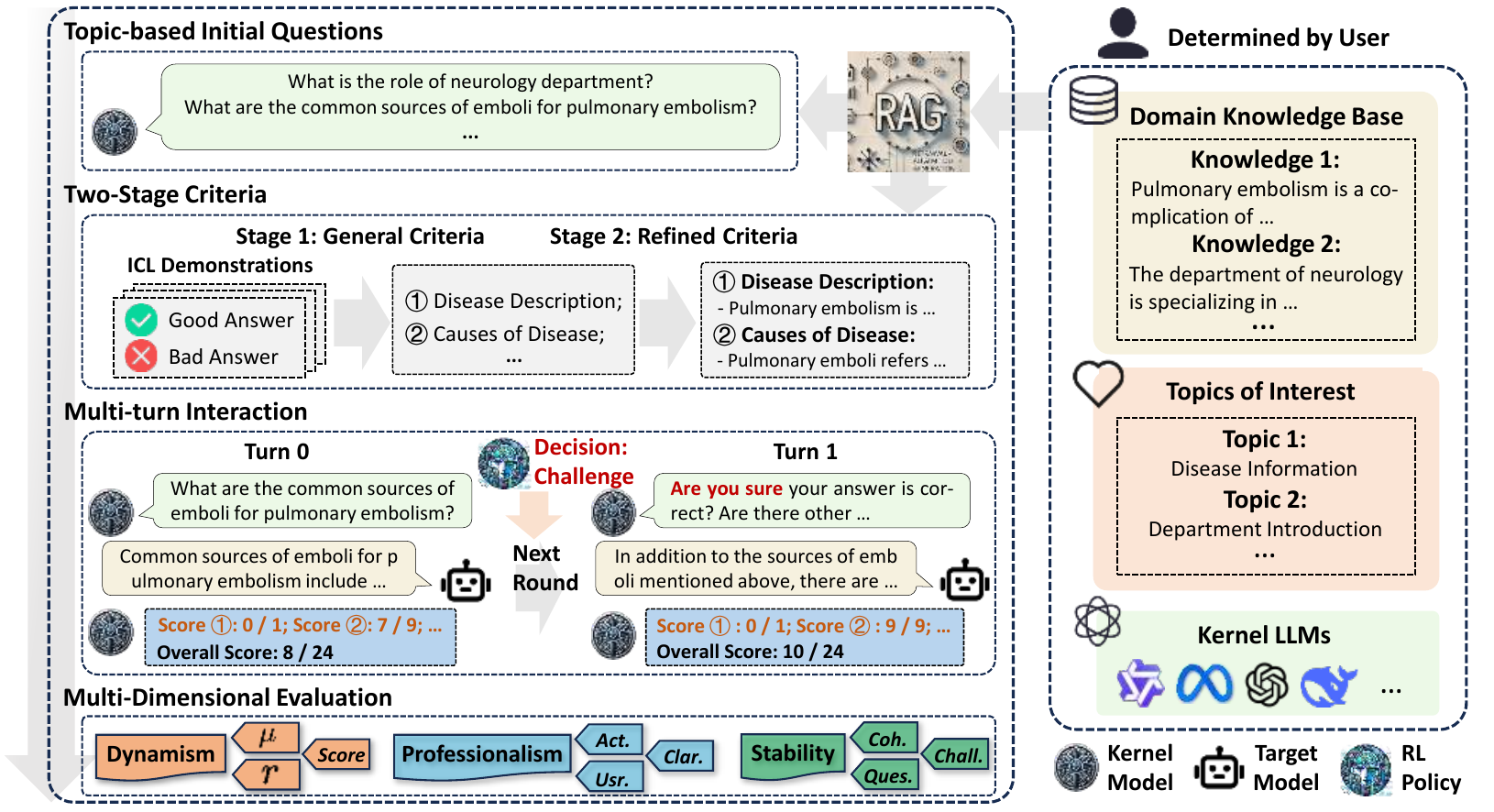}
  \caption{Illustration of \textsc{TestAgent} framework. Leveraging a RAG system built upon the domain knowledge base and topics, it first generates initial questions along with a two-stage criteria construction process, progressing from general topic-level criteria to refined question-specific criteria. Reinforcement Learning then guides the subsequent questioning strategy, determining whether to issue challenges or generate follow-up questions. A multi-dimensional analysis of dynamism, professionalism, and stability is ultimately derived from these interactions.
  }
  \label{fig:frame}
  \vspace{-3pt}
\end{figure*}

\section{Methodology}
\textsc{TestAgent} employs a RAG-based approach to generate questions from vertical domain documents and implements a two-stage criteria generation process from topic-level to question-level, thereby achieving automated benchmark construction. Building upon the generated benchmark, it utilizes RL-driven interaction based on response quality to enable dynamic evaluation of professionalism and stability across multi-turn dialogues. Figure \ref{fig:frame} depicts the overall framework.

\subsection{Automatic Benchmark Generation}
\label{section:benchmark+}
\subsubsection{Topic-based Initial Question Generation}
\label{subsec: Topic}
Knowledge in specialized vertical domains is typically held by users who seek to evaluate a model's performance in their area of expertise. Such knowledge is often structured, making it suitable for constructing domain-specific knowledge bases that Retrieval-Augmented Generation (RAG) can exploit to create evaluation cases. In practice, LLMs in vertical domains are expected to address specific topics, such as ``disease consultation '' or ``department introduction'' in the medical domain. Capturing these expectations is crucial for constructing evaluation cases that reflect real-world user needs. Therefore, we employ RAG to generate domain questions conditioned on topic requirements.

Specifically, given a set of users' topics of interest $T$, we first extract relevant information from the provided domain documents, such as professional textbooks and real-world case records, to construct a domain-specific knowledge base $D$. This knowledge base serves as the retrieval corpus for the RAG system. For each topic in $T$, relevant knowledge is retrieved from $D$ and passed to the evaluation model $K$ (the kernel model), which is prompted to generate an initial question set $Q_0$ within the topic scope. 
These questions incorporate domain topics and adopt a role-play approach, making them more representative of downstream applications. Prompt templates are provided in Appendix~\ref{appendix:prompt}.

For example, the generated questions for the disease topic are:
\textit{``What are the common symptoms of allergic shock?'', 
``What are the main clinical manifestations of SARS?''.}
For the topic of department introduction, the generated questions include:
\textit{``What types of diseases are primarily treated in the Department of Interventional Radiology?'',``How is the Department of Endocrinology connected to other medical departments?''.}

Note that $Q_0$ is the initial question in a multi-turn sequence. The generation of subsequent questions is guided by an RL-learned strategy, which will be elaborated in Section~\ref{subsec: RL}.

\subsubsection{Two-Stage Criteria Generation}
To ensure trustworthy evaluation for the generated questions, we propose a two-stage process for constructing evaluation criteria $C$, progressing from general topic-level to refined, question-specific level criteria. 

\noindent\textbf{Stage 1: General Topic-Level Criteria. }
We first construct general evaluation criteria for each topic $T$ using in-context learning (ICL) with the kernel model $K$. 
For each topic, three demonstration examples—each consisting of a question, a good answer, and a bad answer—are included in the prompt. Based on these examples, $K$ is prompted to summarize the answer pattern and extract essential elements. This procedure yields topic-specific, structured criteria that serve as the foundation for subsequent refinement.
\textit{For instance, as provided in Appendix \ref{appendix:case}, the general criteria for the topic "disease" in the medical domain include: (1) Disease Description; (2) Causes of Disease; ...} 

\noindent\textbf{Stage 2: Refined Question-Level Criteria. }
To obtain fine-grained criteria tailored to each specific question $Q_0$, we further refine the general topic-level criteria by injecting content retrieved via the RAG system. For each question, the RAG system invokes the retriever to obtain relevant documents from $D$, which are then incorporated into the prompt to guide $K$ in filling concrete, fine-grained details evaluation points within the general criteria for the particular question. Eventually, this paradigm develops clear and verifiable criteria for various questions, grounded in the structured topic-general criteria and the traceable retrieved knowledge. 
\textit{For example, as provided in Appendix \ref{appendix:case}, for questions on "allergic shock", the criteria include: (1) Disease Description: Allergic shock is a systemic rapid allergic reaction ...; (2) Causes of Disease: Endocrine substances, food ...}
Prompt templates are provided in Appendix~\ref{appendix:prompt}.

\noindent \textbf{Metric.} Given a generated question and its criteria $C$, we evaluate a model-generated answer $A$ with a score $S$, defined as:
\begin{equation}
S(C,A) = \frac{1}{n} \sum_{i=1}^{n} \mathbbm{1}(c_i, A),
\label{eqn:score}
\end{equation}
where $c_i \in C$ denotes the $i$-th criterion and $n=|C|$, $\mathbbm{1}(x, y)$ equals 1 if $y$ correctly includes $x$ and 0 otherwise. In practice, we adopt the kernel model $K$ to assess the response against each ${c_i}$. 

This hierarchical approach enhances the reliability and controllability of criterion construction, while aligning evaluation goals with user-defined topic intentions. Moreover, through the traceability of retrieved knowledge, the generated questions and criteria remain verifiable. Leveraging structured, source-grounded criteria, the scoring process also ensures objective and consistent evaluation.

It is worth noting that, since knowledge in vertical domains is typically interconnected, the criteria derived from the initial question $Q_0$ encompass a broader set of related knowledge standards through retrieval. Thus, these criteria not only support the evaluation of $Q_0$ but also provide the foundation for generating multiple relevant follow-up questions. 
Consequently, the same criteria are utilized throughout the entire multi-turn interaction sequence initiated by $Q_0$. 

\subsection{Dynamic Multi-turn Interaction}
\label{section:validation+}
\subsubsection{Follow-up Question Generation}
As the criteria generated for an initial question contain comprehensive knowledge relevant to the question through retrieval, it naturally supports the derivation of multiple contextually related follow-up questions. \textit{For example, for the initial question ``What is allergic shock? '', subsequent questions may address its causes, symptoms, or treatments. }Such follow-up generation enables in-depth assessment of domain-specific capability and mirrors realistic user interactions, where users often engage in multi-turn dialogue around a specific topic.

However, fixedly generating follow-up questions is insufficient for exploring the dynamic behaviors of LLMs, such as consistency and stability, which are critical in vertical domains. 
To address this, we propose a dynamic assessment paradigm that leverages feedback from prior responses to adjust the question strategy in real time. 

\subsubsection{RL-based Subsequent Question Decision}
\label{subsec: RL}
To enable exploratory evaluation, the subsequent question should adapt based on the prior response quality of the target model $M$.
This sequential dependency naturally motivates a reinforcement learning formulation, where a policy $\pi_\theta$ determines the subsequent question based on the response quality: $Q_{j+1} := \pi_{\theta}(\cdot)$.
However, treating the raw text questions $Q_{j+1}$ as chosen actions of RL would result in an excessively large action space. To address this, we constrain the action space to a finite set of high-level question types $\mathcal{A} = \{a_1, a_2, \dots, a_m\}$. Then the corresponding prompt of selected type is passed to the kernel model $K$ along with the interaction history and the 
criteria to generate the subsequent concrete question. 
In practice, users tend to pose follow-up questions to $M$ when seeking additional information, and to challenge $M$ for further clarification when responses are incorrect.
To this end, we train $\pi_{\theta}$ to decide whether to ask follow-up questions or challenge based on the answer quality of $M$. The RL components are designed as follows:

\noindent \textbf{State space} includes: \\
1. \textbf{Score and its variation} capture the degree of knowledge acquisition. Therefore, we incorporate the current score, the score variation, and the cosine similarity between consecutive responses. \\
2. \textbf{Last action} is included to obtain the question type and expected performance.

\noindent \textbf{Action space}: follow-up questions ($a_\text{f}$) for seeking additional information and challenging questions ($a_\text{c}$) for further verification.

\noindent \textbf{Reward function} is defined as
\begin{equation}
\begin{cases} 
    1,  \text{if } (|\Delta_t| = 0 \land a_t = a_\text{c}) \lor (|\Delta_t| > c \land a_t = a_\text{f}) \\
    \left|\Delta_t\right| + \alpha(1 - \cos_t), \text{otherwise}
\end{cases}
\label{eq:reward}
\end{equation}
where $\small \Delta_t =  S(C,A'_{t}) - S(C,A'_{t-1})$ is the score difference between summarized current and previous responses $A'_{t} $ and $ A'_{t-1} $; $\cos_t=\cos(\text{emb}(A'_{t+1}), \text{emb}(A'_{t}))$ denotes the cosine similarity between the embedded summarized responses; $c$ is the score fluctuation threshold and $\alpha$ is the semantic variation weight, set as $0.5$ in experiments. The purpose of summarizing the responses is to obtain the gain of the current question-answer interaction. This reward structure encourages the policy to challenge unstable or inconsistent answers and to ask follow-up questions when the model provides substantial new information. 

The RL design leverages real-time feedback and scores to guide multi-turn interactions, exploring the model’s knowledge boundaries while timely challenging the responses to examine stability. Ultimately, the framework identifies the reliable performance boundaries and helps to evaluate professionalism and stability.
We adopt Proximal Policy Optimization (PPO) \cite{ppo} to train the policy, and the complete RL procedure is detailed in Algorithm~\ref{alg:rl} in Appendix~\ref{appendix:rl_alogo}.

\subsection{Multi-Dimensional Evaluation Metrics}
\label{sec:metrics}
Built upon the dynamic multi-turn interactions, we evaluate the target model along three dimensions: Dynamism, Professionalism, and Stability.

\noindent\textbf{Dynamism} quantitatively assesses the model’s performance during multi-turn interactions. \textit{Score} is defined as the cumulative last-turn score $S$ across conversations, representing domain-specific expertise. To capture interaction dynamics, we compute $\mu_\text{f}$, the mean score change after follow-up questions, which reflects the model’s ability to provide additional information when met new questions. We further introduce $r^+$ and $r^-$, the proportions of positive and negative score changes under challenge questions, respectively, together with their mean magnitudes $\mu^+$ and $\mu^-$. Accordingly, $r^+\mu^+$ reflects the model’s corrective capacity, while $r^-\mu^-$ reveals its knowledge instability. Then, the composite metric $\Delta = r^+\mu^+ - r^-\mu^-$ integrates corrective ability and instability, serving as an overall measure of robustness under challenge.

\noindent\textbf{Professionalism} offers a qualitative evaluation of response quality. It considers three aspects: Accuracy (\textit{Act.}), the factual correctness of responses; User satisfaction (\textit{Usr.}), the degree to which responses meet user needs with completeness and contextual relevance; and Clarity (\textit{Clar.}), the readability and structural clarity of responses.

\noindent\textbf{Stability} qualitatively assesses reliability across dynamic interactions, including: Coherence (\textit{Coh.}), the consistency of information across turns; Challenge-handling (\textit{Chall.}), the ability to maintain correctness when confronted with challenges; and Question-handling (\textit{Ques.}), the responsiveness and informativeness of answers to follow-up questions. Both Professionalism and Stability are qualitatively evaluated by GPT-4o.
  
\section{Experiments}
\subsection{Experiment Setup}
\noindent \textbf{Models and Domains.} We evaluate three vertical domains: government, medical, and legal. DeepSeek-V3 \cite{deepseekv2} is employed as the kernel model. Seven target LLMs are evaluated, including GPT-4o \cite{gpt4o}, DeepSeek-V3 \cite{deepseekv2}, LLaMA3.1-70B-Instruct \cite{llama}, LLaMA3.1-405B-Instruct, Ministral-8B \cite{Ministral}, Qwen2.5-7B-Instruct \cite{qwen}, and Qwen2.5-72B-Instruct. Additionally, three medical-specific models (ZhongJing \cite{zhongjing}, BianQue \cite{bianque}, and DoctorGLM \cite{doctorglm}) are compared. The RL policy and value networks are 3-layer MLPs with hidden dimensions $256–128$, trained for $10,000$ steps.

\noindent \textbf{Knowledge Bases.} 
For the medical domain, the knowledge base is built from disease documents in OpenKG~\cite{openkg} and department introductions from CareGPT~\cite{caregpt}, covering disease consultation and department introduction topics. In the legal domain, crime data \cite{crime_type} and judgment data from CAIL \cite{cail2018} support topics of crime knowledge and judgment analysis. For government domain, data from the Chinese Fund Management Center \cite{gjj} is organized into eight fund topics: individual, unit, establishment, amendment, disbursement, loan, repayment, and inquiry. The statistic is shown in Table \ref{tab:benchmark_stats}. 
Raw knowledge is stored in TXT and JSON formats. For retrieval, we combine M3-Embedding \cite{bge_m3} with BM25 \cite{bm25} and adopt BCE-Reranker \cite{bce_reranker}.

\begin{table}[t]
\centering
\resizebox{\linewidth}{!}{
\begin{tabular}{l|c|c|c}
\toprule
\textbf{Domain} & \textbf{Sources} & \textbf{Knowledge Units} & \textbf{Volume} \\
\midrule
\multirow{2}{*}{Legal} 
 & 1000 Legal Cases & 6 & 6000 \\
 & 856 Charges & 6 & 4950 \\
\midrule
\multirow{2}{*}{Medical} 
 & 50 Departments & 7 & 350 \\
 & 808 Diseases & 24 & 19392 \\
\midrule
Government 
 & 102 Affairs & 6 & 612 \\
\bottomrule
\end{tabular}
}
\caption{Sources and capability for question generation. Each source consists of multiple knowledge units. In total, \textsc{TestAgent} can produce 10,950 questions in the legal domain, 19,742 in the medical domain, and 612 in the government domain.}
\label{tab:benchmark_stats}
\end{table}

\subsection{Benchmark Comparison }
Table~\ref{tab:benchmark_compare} compares \textsc{TestAgent} with prevalent LLM benchmarks. Existing benchmarks either rely on static datasets \cite{guha2023legalbench,liu2024medbench,mtbench} or are limited to evaluating general capabilities (e.g., reasoning, comprehension) \cite{dyval,kieval}. Importantly, all methods lack the adaptability of generating benchmarks from scratch across vertical domains. 
Uniquely, \textsc{TestAgent} automates benchmark construction for vertical domains and supports cross-domain generation. Furthermore, it employs exploratory multi-turn interactions that adapt questions to response quality. Overall, it enables vertical domain evaluation and captures nuanced performance through active strategies.

\begin{table}[t]
    \centering
    \resizebox{\linewidth}{!}{
    \begin{tabular}{ccccccc}
    \toprule
         \textbf{Benchmark} & \textbf{Vertical} & \textbf{Adaptable} & \textbf{Dynamic} & \textbf{Multi-turn} & \textbf{Exploration} \\
         \midrule
MT-Bench & \ding{55} & \ding{55} & \ding{55} & \ding{51} & \ding{55} \\
DyVal & \ding{55} & \ding{55} & \ding{51} & \ding{55} & -  \\
KIEval & \ding{55} & \ding{55} & \ding{51} & \ding{51} & \ding{55}  \\
SQuAD & \ding{55} & \ding{55} & \ding{55} & \ding{55} & - \\
LegalBench & \ding{51} & \ding{55} & \ding{55} & \ding{55} & -  \\
MedBench & \ding{51} & \ding{55} & \ding{55} & \ding{55} & - \\
\textbf{TestAgent} & \ding{51} & \ding{51} & \ding{51} & \ding{51} & \ding{51} \\
         \bottomrule
    \end{tabular}
    }
\caption{Comparison of \textsc{TestAgent} with other benchmarks.
Vertical: vertical domains.
Adaptable: cross-domain generation from scratch.
Dynamic: dynamic generation.
Multi-turn: multi-turn evaluation.
Exploration: adjust question strategy based on responses.}
\label{tab:benchmark_compare}
\end{table}

\begin{table}[t]
  \centering
  \resizebox{\linewidth}{!}{
  \begin{tabular}{lccccc}
    \toprule
    & \textbf{Relevancy} & \textbf{Scenarios} & \textbf{Coherence} & \textbf{Completeness} & \textbf{Logic} \\
    \midrule
    Mean      & 4.93 & 4.80 & 4.90 & 4.79 & 4.85 \\
    Variance  & 0.0014 & 0.013 & 0.0035 & 0.014 & 0.0076 \\
    \bottomrule
  \end{tabular}}
  \caption{Human evaluation of generated questions and criteria. Mean and variance for each aspect are reported.}
  \label{tab:Question_criteria}
\end{table}

\subsection{Meta-Evaluation}
We conduct meta-evaluations to assess the quality of generated questions and criteria, as well as the reliability of kernel-model scoring.

For the generated questions, human experts rated three aspects on a 1–5 scale: \textit{Relevancy} (domain appropriateness and alignment with the target topic), \textit{Scenarios} (grounding in realistic user contexts), and \textit{Coherence} (topic consistency across multiple dialogue turns).
For the criteria, two aspects were evaluated: \textit{Completeness} (coverage of key aspects required to answer the questions) and \textit{Logic} (soundness of structure and reasoning).
As shown in Table~\ref{tab:Question_criteria}, both questions and criteria received consistently high scores with minimal variance, indicating strong agreement with human judgment.  

To assess kernel-model scoring, experts indicated agreement (1) or disagreement (0) with each score. GPT-4o achieved a mean agreement of $\mu=0.974$ with variance $\sigma^2=0.025$, while DeepSeek-V3 achieved $\mu=0.957$ with $\sigma^2=0.040$. Although GPT-4o performed slightly better, both models exhibited high reliability, confirming the validity of the scoring process.
Taken together, these findings demonstrate the robustness of \textsc{TestAgent} in generating high-quality evaluation data and ensuring reliable scoring.

\begin{table*}[t]
\setlength{\tabcolsep}{4.5pt}
\centering
\scriptsize
\begin{tabular}{
>{\raggedright\arraybackslash}p{0.25cm}|
>{\raggedright\arraybackslash}p{1.5cm}|
>{\centering\arraybackslash}p{0.6cm}
>{\centering\arraybackslash}p{0.6cm}
>{\centering\arraybackslash}p{0.6cm}
>{\centering\arraybackslash}p{0.6cm}
>{\centering\arraybackslash}p{0.6cm}
>{\centering\arraybackslash}p{0.6cm}|
>{\centering\arraybackslash}p{0.7cm}
>{\centering\arraybackslash}p{0.7cm}
>{\centering\arraybackslash}p{0.7cm}|
>{\centering\arraybackslash}p{0.3cm}
>{\centering\arraybackslash}p{0.3cm}
>{\centering\arraybackslash}p{0.3cm}
>{\centering\arraybackslash}p{0.3cm}|
>{\centering\arraybackslash}p{0.3cm}
>{\centering\arraybackslash}p{0.3cm}
>{\centering\arraybackslash}p{0.3cm}
>{\centering\arraybackslash}p{0.3cm}}
\toprule
~ &\multicolumn{1}{l|}{\multirow{2}{*}{\textbf{Models}}} &
\multicolumn{9}{c|}{\textbf{Dynamism}} &
\multicolumn{4}{c|}{\textbf{Professionalism} $\uparrow$} &
\multicolumn{4}{c}{\textbf{Stability} $\uparrow$} \\
~ & ~ &\textit{Score}$\uparrow$ & $\mu_\text{f}$$\uparrow$ & $r^+$$\uparrow$ & $\mu^+$$\uparrow$ & $r^-$$\downarrow$& $\mu^-$$\downarrow$ & $r^+\mu^+$$\uparrow$ & $r^-\mu^-$$\downarrow$ & $\Delta$$\uparrow$
& \textit{Act.} & \textit{Usr.} & \textit{Clar.} & \textit{Avg.}
& \textit{Coh.} & \textit{Ques.} & \textit{Chall.} & \textit{Avg.} \\
\midrule
\multirow{10}{*}{\rotatebox{90}{\textbf{Medical}}}  &\texttt{GPT-4o} & \textbf{67.54} & \textbf{0.121} & \underline{0.803} & \textbf{0.042} & \underline{0.197} & 0.086 & \textbf{0.034} & 0.017 & \underline{0.017} & 7 & 8 & 8 & \textbf{7.7} & 7 & 8 & 8 & \underline{7.7} \\
~ &\texttt{DeepSeek-V3} & 65.67 & 0.115 & \textbf{0.823} & 0.034 & \textbf{0.177} & 0.082 & 0.028 & 0.015 & 0.013 & 7 & 8 & 8 & \textbf{7.7} & 7 & 7 & 7 & 7.0 \\
~ &\texttt{Qwen2.5-7B} & 61.01 & 0.113 & 0.802 & 0.040 & 0.198 & 0.086 & 0.032 & 0.017 & 0.015 & 6 & 6 & 8 & 6.7 & 7 & 7 & 7 & 7.0 \\
~ &\texttt{Qwen2.5-72B} & \underline{66.95} & \underline{0.119} & 0.802 & \underline{0.041} & 0.198 & \textbf{0.058} & \underline{0.033} & \textbf{0.011} & \textbf{0.021} & 7 & 7 & 8 & \underline{7.3} & 8 & 8 & 8 & \textbf{8.0} \\
~ &\texttt{LLaMA3.1-70B} & 61.51 & 0.098 & 0.812 & 0.033 & 0.188 & \underline{0.064} & 0.027 & \underline{0.012} & 0.015 & 6 & 6 & 8 & 6.7 & 8 & 7 & 5 & 6.7 \\
~ &\texttt{LLaMA3.1-405B} & 61.54 & 0.104 & 0.781 & 0.026 & 0.219 & 0.083 & 0.020 & 0.018 & 0.002 & 6 & 7 & 8 & 7.0 & 7 & 7 & 8 & 7.3\\
~ &\texttt{Ministral-8B} & 64.70 & 0.108 & 0.819 & 0.039 & 0.181 & 0.083 & 0.032 & 0.017 & 0.015 & 6 & 7 & 8 & 7.0 & 7 & 7 & 7 & 7.0 \\
~ &\cellcolor{gray!10}\texttt{ZhongJing} &\cellcolor{gray!10} 41.45 &\cellcolor{gray!10} 0.099 &\cellcolor{gray!10} 0.713 &\cellcolor{gray!10} 0.022 &\cellcolor{gray!10} 0.287 &\cellcolor{gray!10} 0.087 &\cellcolor{gray!10} 0.016 &\cellcolor{gray!10} 0.025 &\cellcolor{gray!10} -0.009 &\cellcolor{gray!10} 5 &\cellcolor{gray!10} 5 &\cellcolor{gray!10} 6 &\cellcolor{gray!10} 5.3 &\cellcolor{gray!10} 5 &\cellcolor{gray!10} 6 &\cellcolor{gray!10} 4 &\cellcolor{gray!10} 5.0 \\
~ &\cellcolor{gray!10}\texttt{BianQue} &\cellcolor{gray!10} 30.19 &\cellcolor{gray!10} 0.088 &\cellcolor{gray!10} 0.660 &\cellcolor{gray!10} 0.021 &\cellcolor{gray!10} 0.340 &\cellcolor{gray!10} 0.084 &\cellcolor{gray!10} 0.014 &\cellcolor{gray!10} 0.029 &\cellcolor{gray!10} -0.015 &\cellcolor{gray!10} 3 &\cellcolor{gray!10} 4 &\cellcolor{gray!10} 5 &\cellcolor{gray!10} 4.0 &\cellcolor{gray!10} 4 &\cellcolor{gray!10} 4 &\cellcolor{gray!10} 3 &\cellcolor{gray!10} 3.7 \\
~ &\cellcolor{gray!10}\texttt{DoctorGLM} &\cellcolor{gray!10} 34.17 &\cellcolor{gray!10} 0.075 &\cellcolor{gray!10} 0.663 &\cellcolor{gray!10} 0.025 &\cellcolor{gray!10} 0.337 &\cellcolor{gray!10} 0.083 &\cellcolor{gray!10} 0.017 &\cellcolor{gray!10} 0.028 &\cellcolor{gray!10} -0.011 &\cellcolor{gray!10} 3 &\cellcolor{gray!10} 4 &\cellcolor{gray!10} 6 &\cellcolor{gray!10} 4.3 &\cellcolor{gray!10} 4 &\cellcolor{gray!10} 3 & \cellcolor{gray!10} 3 &\cellcolor{gray!10} 3.3 \\
\midrule
\multirow{7}{*}{\rotatebox{90}{\textbf{Legal}}} &\texttt{GPT-4o} & \textbf{70.99} & 0.145 & \underline{0.827} & \textbf{0.100} & \underline{0.173} & \textbf{0.125} & \textbf{0.083} & \underline{0.022} & \underline{0.061} & 7 & 7 & 8 & \textbf{7.3} & 6 & 8 & 8 & \underline{7.3} \\
~ &\texttt{DeepSeek-V3} & 68.28 & 0.146 & \textbf{0.884} & 0.091 & \textbf{0.116} & 0.134 & \underline{0.080} & \textbf{0.016} & \textbf{0.065} & 6 & 7 & 8 & \underline{7.0} & 7 & 8 & 8 & \textbf{7.7} \\
~ &\texttt{Qwen2.5-7B} & 64.96 & \textbf{0.164} & 0.772 & 0.087 & 0.228 & 0.137 & 0.067 & 0.031 & 0.036 & 6 & 6 & 7 & 6.3 & 5 & 8 & 7 & 6.7 \\
~ &\texttt{Qwen2.5-72B} & \underline{70.34} & \underline{0.152} & 0.773 & \underline{0.098} & 0.227 & 0.132 & 0.076 & 0.030 & 0.046 & 7 & 7 & 8 & \textbf{7.3} & 6 & 8 & 7 & 7.0 \\
~ &\texttt{LLaMA3.1-70B} & 59.50 & 0.119 & \underline{0.827} & 0.083 & \underline{0.173} & 0.132 & 0.069 & 0.023 & 0.046 & 5 & 6 & 7 & 6.0 & 6 & 6 & 7 & 6.3 \\
~ &\texttt{LLaMA3.1-405B} & 55.97 & 0.131 & 0.810 & 0.085 & 0.190 & 0.142 & 0.069 & 0.027 & 0.042 & 4 & 5 & 6 & 5.0 & 6 & 7 & 7 & 6.7 \\
~ &\texttt{Ministral-8B} & 61.79 & 0.140 & 0.755 & 0.094 & 0.245 & \underline{0.128} & 0.071 & 0.031 & 0.040 & 5 & 6 & 7 & 6.0 & 5 & 7 & 6 & 6.0 \\
\midrule
\multirow{7}{*}{\rotatebox{90}{\textbf{Government}}} & \texttt{GPT-4o} & \textbf{57.84} & 0.092 & \underline{0.844} & \textbf{0.051} & 0.156 & 0.079 & \textbf{0.043} & 0.012 & \underline{0.031} & 6 & 7 & 8 & \textbf{7.0} & 7 & 7 & 8 & \textbf{7.3} \\
~ & \texttt{DeepSeek-V3} & 52.73 & \underline{0.107} & 0.800 & \textbf{0.051} & 0.200 & \underline{0.069} & \underline{0.041} & 0.014 & 0.027 & 6 & 7 & 7 & \underline{6.7} & 6 & 8 & 7 & \underline{7.0} \\
~ &\texttt{Qwen2.5-7B} & 48.27 & 0.088 & 0.836 & 0.042 & 0.164 & 0.075 & 0.035 & 0.012 & 0.013 & 5 & 6 & 7 & 6.0 & 7 & 6 & 6 & 6.3 \\
~ &\texttt{Qwen2.5-72B} & \underline{57.38} & \textbf{0.108} & \textbf{0.858} & 0.044 & \underline{0.142} & 0.080 & 0.038 & \underline{0.011} & 0.026 & 6 & 7 & 8 & \textbf{7.0} & 7 & 8 & 7 & \textbf{7.3} \\
~ &\texttt{LLaMA3.1-70B} & 49.10 & 0.081 & 0.849 & \underline{0.048} & 0.151 & \textbf{0.056} & \underline{0.041} & \textbf{0.008} & \textbf{0.032} & 5 & 6 & 7 & 6.0 & 8 & 6 & 8 & \textbf{7.3} \\
~ &\texttt{LLaMA3.1-405B} & 46.77 & 0.080 & 0.840 & 0.040 & 0.160 & 0.080 & 0.034 & 0.013 & 0.021 & 5 & 6 & 7 & 6.0 & 6 & 6 & 7 & 6.3 \\
~ &\texttt{Ministral-8B} & 48.30 & 0.088 & 0.843 & 0.043 & 0.157 & 0.087 & 0.036 & 0.014 & 0.023 & 5 & 5 & 7 & 5.7 & 6 & 6 & 7 & 6.3 \\
\bottomrule
\end{tabular}
\caption{
Evaluation results on the medical, legal, and government domains. 
Bold and underlined values denote the best and second best. 
\textit{Score} ranges from 0–100, while Professionalism and Stability are rated on a 1–10 scale. 
Domain-specific models are shaded in gray.}
\label{tab:main}
\vspace{-3pt}
\end{table*}

\subsection{Main Results}
We conducted 100 conversations per domain, with 6 turns per conversation, yielding 600 questions per domain. Main evaluation results, covering dynamism, professionalism, and stability, are presented in Table~\ref{tab:main}. Topic-specific scores are shown in Figure \ref{fig:heatmap}.

\begin{table*}[t]
\begin{minipage}[c]{0.65\textwidth}
    \centering 
   \setlength{\tabcolsep}{3.8pt}
\scriptsize
\begin{tabular}{
>{\raggedright\arraybackslash}p{1.8cm}|
>{\centering\arraybackslash}p{0.6cm}
>{\centering\arraybackslash}p{0.6cm}
>{\centering\arraybackslash}p{0.6cm}
>{\centering\arraybackslash}p{0.6cm}
>{\centering\arraybackslash}p{0.6cm}
>{\centering\arraybackslash}p{0.6cm}|
>{\centering\arraybackslash}p{0.7cm}
>{\centering\arraybackslash}p{0.7cm}
>{\centering\arraybackslash}p{0.7cm}}
\toprule
\textbf{Models} & \textit{Score}$\uparrow$ & $\mu_{\text{f}}$$\uparrow$ & $r^+$$\uparrow$ & $\mu^+$$\uparrow$ & $r^-$$\downarrow$ & $\mu^-$$\downarrow$ &
$r^+\mu^+$$\uparrow$ & $r^-\mu^-$$\downarrow$ & $\Delta$$\uparrow$ \\
\midrule
GPT-4o & 99.10 & 0.119 & 0.892 & 0.092 & 0.066 & 0.088 & 0.0821 & 0.0058 & 0.0763 \\
DeepSeek-V3 & \textbf{99.57} & \textbf{0.121} & \textbf{0.905} & 0.093 & \textbf{0.045} & \textbf{0.085} & \textbf{0.0842} & \textbf{0.0038} & \textbf{0.0804} \\
Qwen2.5-7B & 98.66 & 0.117 & 0.874 & \textbf{0.094} & 0.055 & 0.086 & 0.0822 & 0.0047 & 0.0775 \\
Qwen2.5-72B & 98.82 & 0.121 & 0.891 & 0.091 & 0.063 & \textbf{0.085} & 0.0811 & 0.0054 & 0.0757 \\
LLaMA3.1-70B & 97.69 & 0.118 & 0.851 & 0.092 & 0.095 & 0.087 & 0.0783 & 0.0083 & 0.0700 \\
LLaMA3.1-405B & 98.87 & 0.118 & 0.859 & 0.093 & 0.093 & 0.089 & 0.0799 & 0.0083 & 0.0716 \\
Ministral-8B & 98.86 & 0.115 & 0.797 & 0.093 & 0.143 & 0.086 & 0.0741 & 0.0123 & 0.0618 \\
\bottomrule
\end{tabular}
\caption{Evaluation results in reading comprehension domains, which adapt \textsc{TestAgent} to activate the existing static benchmark.}
\label{tab:general}

\end{minipage}%
\hfill
\begin{minipage}[c]{0.33\textwidth}
    \centering 
    \raisebox{7pt}{%
    \includegraphics[width=\linewidth]{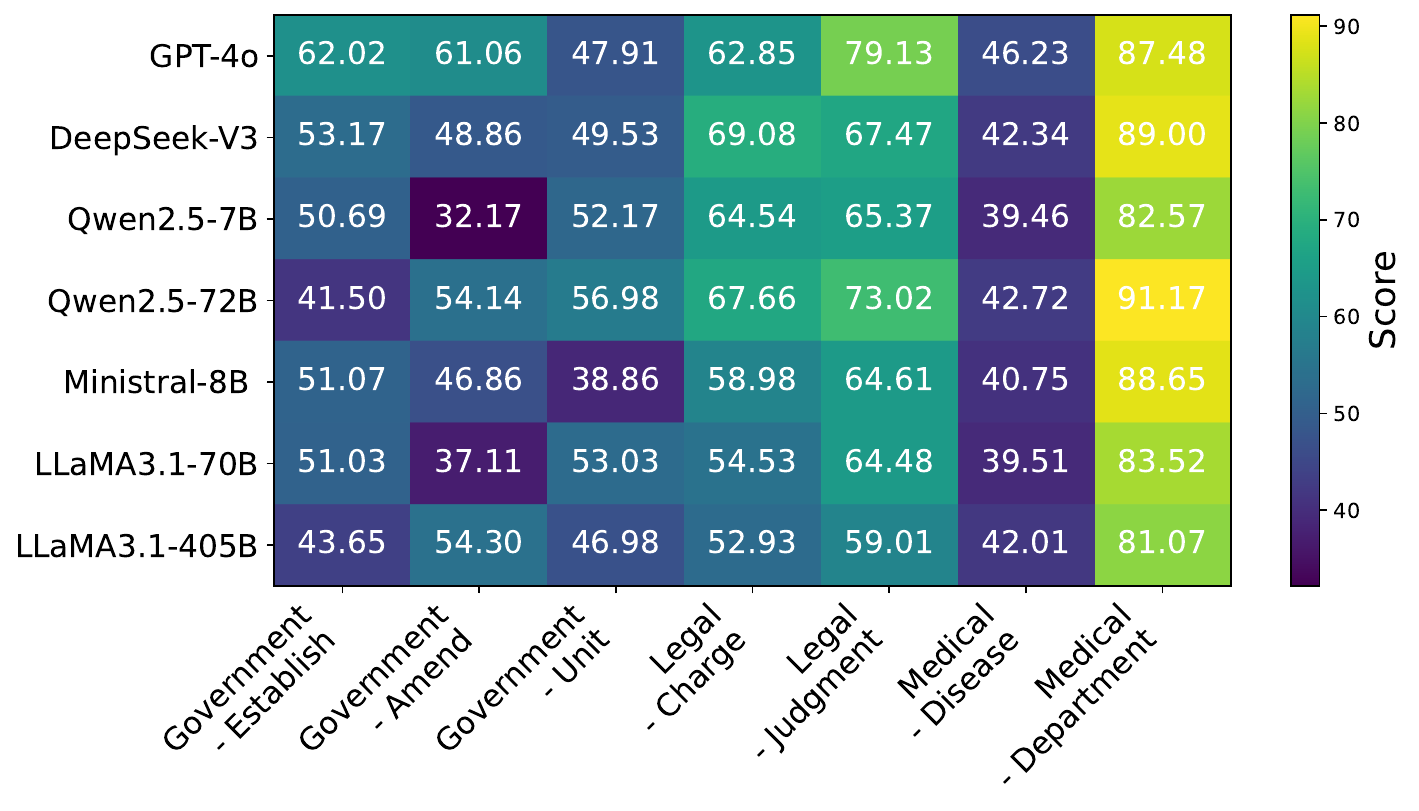} 
    }
    \captionsetup{skip=1pt}  
    \caption{\textit{Scores} on specific topics of vertical domains.} 
    \label{fig:heatmap}
\end{minipage}
\vspace{-5pt}
\end{table*}

\begin{figure*}
    \centering
    \includegraphics[width=\linewidth]{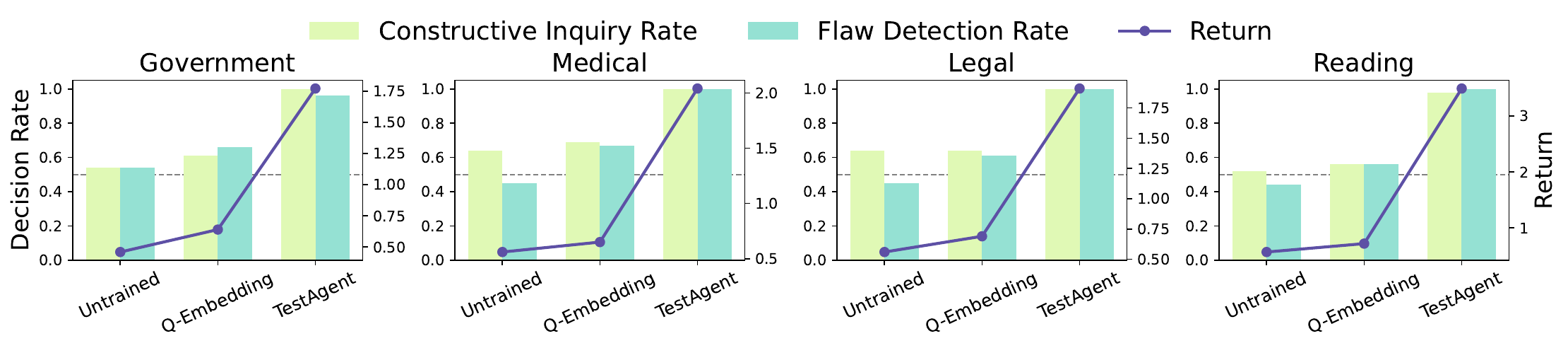}
    \caption{Strategic evaluation performance across domains. Comparison of \textsc{TestAgent} with Untrained baseline and Q-Embedding baseline, where the RL state space incorporates question embeddings. 
    The dashed horizontal line at 0.5 represents random selection.
    }
 \label{fig:ratio}
 \vspace{-6pt}
\end{figure*}

\begin{figure}[t]
    \centering
    \includegraphics[width=\linewidth]
    {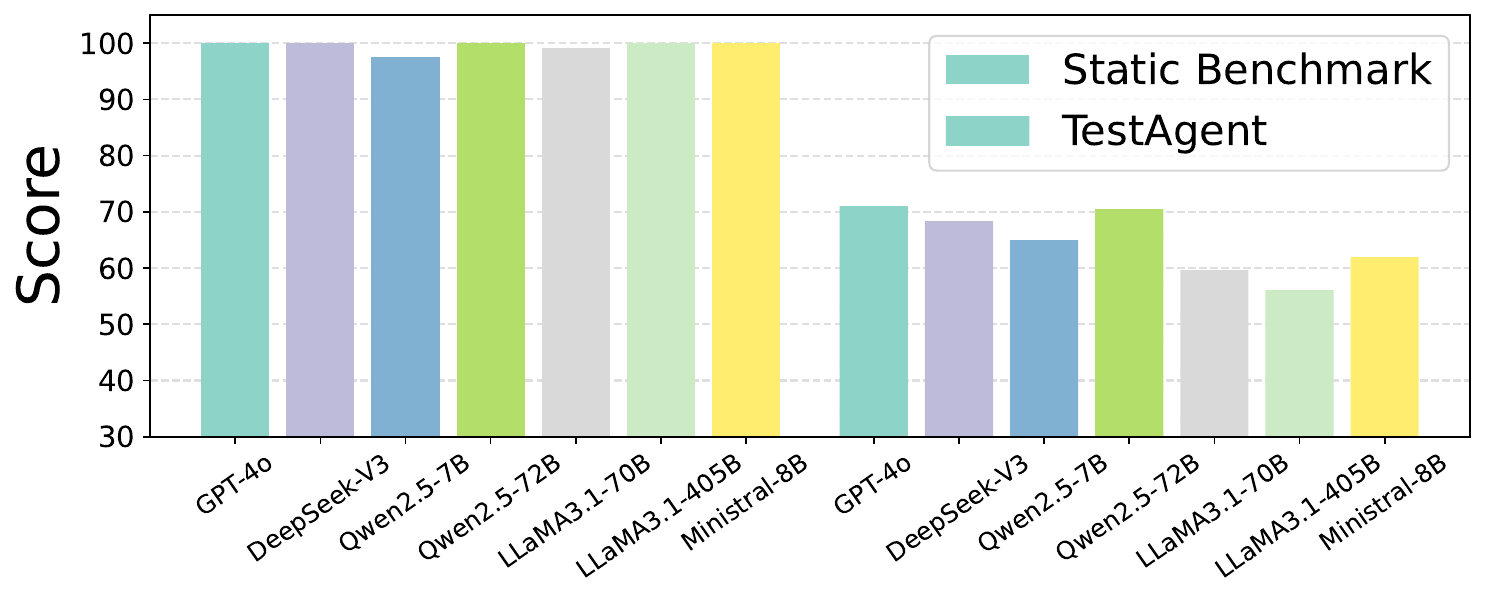} 
\captionof{figure}{Static single-turn evaluation versus \textsc{TestAgent} dynamic evaluation in the legal domain.} 
      \label{fig:static_benchmark_vs_testagent}
\end{figure}

\noindent\textbf{Medical Domain.} 
GPT-4o demonstrates a relatively high level of medical expertise and knowledge provision, achieving both the top \textit{Score} and score gain $\mu_f$, followed closely by Qwen2.5-72B-Instruct and DeepSeek-V3. Dynamic metrics further distinguish top models: while GPT-4o excels at error correction (highest $r^+\mu^+$), Qwen2.5-72B-Instruct demonstrates superior stability under challenge (lowest $r^-\mu^-$ and highest $\Delta$). 
In contrast, BianQue, ZhongJing, and DoctorGLM perform poorly, with low \textit{Score} and $\mu_\text{f}$, as well as negative $\Delta$, revealing limited knowledge reserves and difficulty addressing novel questions. Moreover, these models tend to doubt their own answers and yield under challenge.
Qualitative evaluation of \textit{Professionalism} and \textit{Stability} align these findings: GPT-4o and DeepSeek-V3 excel in factual accuracy, user satisfaction, and clarity, while Qwen2.5-72B-Instruct leads in stability by effectively handling diverse question types.
By contrast, the three domain-specific models show significant deficiencies. Although they are fine-tuned for the medical domain, weaknesses in their base models and the narrow scope of training contribute to poor performance in realistic multi-turn scenarios.

\noindent\textbf{Legal Domain.} 
Similar trends are observed in the legal domain: GPT-4o ranks highest, followed by Qwen2.5-72B-Instruct and DeepSeek-V3. Notably, DeepSeek-V3 achieves the highest proportion of rational analysis and error correction, as indicated by its elevated $r^+$, while its low $r^-\mu^-$ reflects strong knowledge stability. The high \textit{Coh.} further corroborates its ability to maintain consistency across turns. Interestingly, when responding to follow-up questions, the smaller Qwen2.5-7B-Instruct demonstrates strong interaction efficiency, reflected in its high $\mu_f$. However, due to its poor stability, which manifests in greater response variability and an increased tendency to produce contradictory answers, its overall performance remains relatively low.

\noindent\textbf{Government Domain.} In this domain, 
GPT-4o and Qwen2.5-72B-Instruct achieve comparable top \textit{Scores}, with GPT-4o excelling in error correction (high $r^+$, $r^+\mu^+$), while Qwen2.5-72B-Instruct performs better in handling follow-up questions (high $\mu_f$ and \textit{Ques.}), suggesting complementary strengths. Notably, none of the models surpass a \textit{Score} of 60, underscoring the difficulty of this highly specialized domain. This highlights the necessity of domain-specific evaluation, as strong general-domain performance does not guarantee proficiency in vertical domains.

\subsection{Activating Existing Benchmarks}
\textsc{TESTAGENT} can introduce dynamism into static benchmarks with minimal modification. Taking the classic reading comprehension benchmark SQuAD~\cite{squad} as an example, the dataset provides multiple independent questions with corresponding answers, grounded in given context paragraphs. Within this setting, the RL policy decides whether ask a follow-up question within the benchmark or challenge the model's answer, with rewards derived from alignment with ground-truth answers. 
On the transformed dynamic benchmark, DeepSeek-V3 attains the highest level of expertise and demonstrates exceptional stability. In contrast to vertical domains, all models achieve \textit{Scores} exceeding 95 on reading comprehension, underscoring their effectiveness in general, non-domain-specific tasks. Importantly, \textsc{TESTAGENT} converts static datasets into dynamic multi-turn interactions, thereby enabling the use of metrics such as $\mu_f$, $r^{+(-)}$ and $\mu^{+(-)}$ to reveal the underlying factors behind high scores, specifically the models’ capacity for multi-turn contextual understanding and knowledge stability.

\subsection{Effect of Multi-turn Evaluation}
As shown in Figure~\ref{fig:static_benchmark_vs_testagent}, we compare static single-turn evaluation with dynamic multi-turn evaluation on the judgment analysis task. Single-turn static questions mainly assess surface-level understanding of case documents and cannot capture the depth and complexity of real-world legal scenarios that require multiple conversational turns, thus suffering from a ceiling effect with minimal performance differences across models.
In contrast, \textsc{TestAgent} integrates multi-turn professional dialogue and considers dynamic consistency, aligning closely with practical evaluation needs. It exposes substantial performance gaps between models, achieving $4.3\times$ greater discriminative power than the static one.

\subsection{RL Improves Evaluation Quality}
To assess the intelligence of evaluation methods, we introduce two strategic metrics: \textit{Flaw Detection Rate}, the frequency of issuing challenges when the model performs poorly (stagnant scores), thereby probing uncertain knowledge regions; and \textit{Exploratory Inquiry Rate}, the frequency of posing follow-up questions when the model exhibits strong performance (single-turn score gain $>0.2$), thereby broadening knowledge exploration. High values on both metrics indicate timely and adaptive questioning, awarding reliable answers while enabling credible and in-depth evaluation.

As shown in Figure \ref{fig:ratio}, higher Returns correspond to higher Flaw Detection and Exploratory Inquiry Rates, demonstrating that RL training improves evaluation quality. Specifically, the Untrained baseline exhibits near-random behavior (Rates $\approx 0.5$), while the Q-Embedding variant fails to learn effective strategies due to limited policy capacity. In contrast, \textsc{TestAgent}, trained with the RL design in Section \ref{subsec: RL}, achieves high Returns and consistently attains near-perfect Flaw Detection and Exploratory Inquiry Rates ($\approx 1$). 
We also test a direct strategy where the LLM determines question types, but it consistently selects follow-ups regardless of response quality.
Overall, these results demonstrate that RL training enables \textsc{TestAgent} to both identify weaknesses in target models and extend evaluations through adaptive questioning, thereby supporting reliable assessment of stable knowledge boundaries.

\section{Conclusion}
In this paper, we propose \textsc{TestAgent}, a framework for evaluating LLMs in vertical domains. By leveraging RAG, \textsc{TestAgent} automatically constructs benchmarks from scratch across multiple domains. It adopts an RL-based exploratory multi-turn interactive approach to dynamically select questioning strategies according to model performance, thereby capturing the dynamic behavior of LLMs and providing a comprehensive assessment of professionalism and stability. Experiments demonstrate the versatility and effectiveness of the framework. \textsc{TestAgent} facilitates efficient benchmark construction and offers valuable insights into LLM performance in vertical domain evaluation.

\bibliography{custom}

\appendix
\onecolumn 

\section{RL Algorithm}
\label{appendix:rl_alogo}
Algorithm~\ref{alg:rl} details the reinforcement learning (RL) procedure implemented in \textsc{TestAgent}. It iteratively improves the question generation strategy to explore the target model’s performance more effectively.
\begin{algorithm}[ht]
\caption{RL in TestAgent}
\textbf{Input}: kernel model $K$, target model $M$, knowledge base $D$, topic $T$, initial question prompt $p_\text{i}$, criterion prompt $p_\text{C}$, score prompt $p_\text{s}$, follow-up question prompt $p_\text{f}$, challenge prompt $p_\text{c}$, threshold $c$, update frequency $n$, max steps $N$, turns per episode $m$\\
Initialize policy $\pi_{\theta}$ randomly, step $=0$, empty buffer $B$.

\begin{algorithmic}[1]
\WHILE{step $< N$}
    \STATE Generate initial question $Q_0 = K(p_\text{i};D,T)$ and criteria $C = K(p_\text{C},Q_0;D,T)$
    \STATE Get answer $A_0 = M(Q_{0})$ and its score $S(C, A_0) = K(p_\text{s},C,A_0)$
    \STATE Set initial state $s_0 = (S(C, A_{0}),S(C, A_{0}),S(C, A_{0}))$ and interaction history $H_0 = \{(Q_0, A_0, S(C, A_0))\}$
    \FOR{turn $t=0,\cdots,m$}
        \STATE Sample action $a_t \sim \pi_{\theta}(s_t)$
        \STATE Generate $Q_{t+1}$ based on $a_t$: Follow-up question 
 $K(p_\text{f},H_t;C)$ if $a_t=a_\text{f}$; Challenge 
 $K(p_\text{c},H_t;C)$ otherwise
        \STATE Get answer $A_{t+1} = M(Q_{t+1})$ and summarize to $A'_{t+1}$ based on $A_0,\cdots,A_{t+1}$
        
        \STATE Score $S(C, A_{t+1}) = K(p_\text{s},C,A'_{t+1})$
        \STATE Update $H_{t+1} = H_{t} \cup \{(Q_{t+1}, A_{t+1}, S(C, A_{t+1}))\}$
        \STATE Compute $\Delta_t = S(C, A'_{t+1}) - S(C, A'_{t})$, $cos_t = \text{cos}(\text{emb}(A'_{t+1}), \text{emb}(A'_{t}))$
        \STATE Set reward $r_t$ via \eqref{eq:reward}, state $s_{t+1} = (S(C, A'_{t+1}), \Delta_t, cos_t)$
        \STATE Add $(s_{t}, a_t, r_t, s_{t+1})$ to $B$
        \STATE $\text{step} \gets \text{step} + 1$
        \IF{step $\%~n = 0$}
            \STATE Update $\pi_{\theta}$ with PPO loss, clear $B$
        \ENDIF
    \ENDFOR
\ENDWHILE
\STATE \textbf{Evaluate} $M$ using trained policy $\pi_{\theta}$
\end{algorithmic}
\label{alg:rl}
\end{algorithm}

\section{Related Work}
\noindent \textbf{Multi-turn Evaluation.} 
Multi-turn evaluation benchmarks have evolved significantly to assess language models' interactive capabilities. 
MT-bench \cite{mtbench} is the first benchmark built for multi-turn instruction-following evaluation, which contains two-turn questions per dialogue. MT-bench-101 \cite{mtbench101} formulated a three-tier hierarchical ability taxonomy and used GPT-4o to generate questions focused on these abilities. Similarly, MT-Eval \cite{mteval} leverages existing datasets augmented with GPT-4o-generated examples. The study by \cite{user} collected authentic feedback from real users, forming a multi-cultural, multi-LLMs user-reported scenario dataset. 

However, these benchmarks predominantly rely on static datasets, thus neglecting the potential advantages of dynamically generated evaluation questions tailored adaptively based on model responses.
 
\noindent \textbf{Dynamic Evaluation.} 
To overcome static benchmark limitations, recent studies have proposed dynamic evaluation frameworks. DyVal \cite{dyval} utilizes directed acyclic graphs to dynamically generate datasets for reasoning tasks. KIEval \cite{kieval} accesses the output of evaluated models to existing datasets and repeatedly confirms the reasons behind the model's choices, thus mitigating data contamination. AgentBench \cite{agentbench} evaluates the ability of LLMs to act as agents in open-ended generation settings. 
Additionally, MINT~\cite{mint} evaluates models' abilities to solve tasks through multi-turn interactions involving external tools and natural language feedback.

Despite these advances, existing dynamic evaluation frameworks primarily focus on general LLM capabilities, posing significant challenges when applied to vertical-specific domains. Domain-specific benchmarks, such as LegalBench~\cite{guha2023legalbench} for legal contexts and MedBench~\cite{liu2024medbench} for medical applications, still rely heavily on manual data collection, thereby limiting scalability, adaptability, and dynamic multi-turn dialogue generation capabilities. 
Most importantly, all existing evaluation frameworks, whether for general or vertical domains, lack cross-domain adaptability, i.e., they are unable to automatically generate benchmarks from scratch across different domains.

To bridge this gap, we propose a novel evaluation framework that automates benchmark construction tailored to arbitrary vertical domains. By revisiting the elements of evaluation, our approach dynamically constructs multi-turn dialogues interactively based on real-time model responses. This enables scalable, adaptive, and fully automated benchmarking, addressing critical limitations inherent in both static benchmarks and existing dynamic evaluation frameworks.

\noindent \textbf{RL for LLMs.} 
Reinforcement learning (RL) has emerged as a promising approach for optimizing prompts and interactions with LLMs. 
RLPROMPT \cite{rlprompt} presents an efficient approach to optimize discrete prompts without the need for gradient information. TEMPERA \cite{tempera} introduces a novel action space for flexible prompt editing, adapting to different queries.
PILLOW \cite{pillow} utilizes RL to enhance the performance of Low-Rank Adaptation by intelligently selecting and concatenating prompts from a user-defined pool, reducing computational costs. PRewrite \cite{prewrite} automates the rewriting of prompts via RL for improved performance on downstream tasks. 
Prompt-OIRL \cite{PromptOIRL} utilizes offline inverse reinforcement learning to optimize prompts economically, demonstrating the potential of RL in prompt engineering for LLMs. 

Drawing inspiration from these approaches, our framework conceptualizes user-centered benchmark construction as a question-challenge decision problem, utilizing RL to perform automatic prompt selection in multi-turn question generation. This RL-driven methodology enables exploration of domain model capabilities through dynamically generated follow-up questions or challenges based on model responses.

\section{Industry Survey Subjects}
\label{appendix:evaluation subjects}
The LLMs industry chain is divided into three layers: the infrastructure layer, the model and tools layer, and the application layer. Our survey covers 16 companies in the infrastructure layer, with leading enterprises including Shanghai Supercomputer Center, Tencent Yangtze River Delta Artificial Intelligence Advanced Computing Center, Alibaba Cloud East China Intelligent Computing Center, and Huawei Software Technologies Co., Ltd. The model and tools layer includes 20 companies, with key players such as Shanghai SenseTime Intelligent Technology Co., Ltd., Shanghai Artificial Intelligence Laboratory, Shanghai Xiyu Technology Co., Ltd., and Huawei Software Technologies Co., Ltd. The application layer consists of 34 companies, with leading enterprises including Transwarp Technology (Shanghai) Co., Ltd., iFLYTEK Co., Ltd., Shanghai Kuanyu Digital Technology Co., Ltd., Alibaba Network Technology Co., Ltd., and Silicon Intelligence Technology Co., Ltd.

\section{Case Study}
\label{appendix:case}
The following cases illustrate the evaluation process in the medical domain, demonstrating that \textsc{TestAgent} can effectively generate benchmarks for vertical domains and perform automatic assessment that meets realistic needs.\\

\noindent\textbf{Initial Questions for Disease Topic}
\begin{tcolorbox}[colback=gray!10, colframe=gray!100, rounded corners, leftrule={0pt}, rightrule={0pt}, toprule={0pt}, bottomrule={0pt}, left={10pt}, right={10pt}, top={5pt}, bottom={5pt},breakable]
1. What are the common symptoms of decompression sickness?\\
2. What are the main clinical manifestations of SARS?\\
3. What are the emergency treatment methods for allergic shock?\\
4. What are the arterial blood gas analysis criteria for respiratory failure?\\
5. What are the common symptoms of tuberculous empyema?\\
6. How can the occurrence of septic shock be prevented?\\
7. What are the main causes of sick building syndrome?\\
8. How can the occurrence of respiratory foreign bodies in children be prevented?
\end{tcolorbox}

\noindent\textbf{General Criterion for Disease Topic}
\begin{tcolorbox}[colback=gray!10, colframe=gray!100, rounded corners, leftrule={0pt}, rightrule={0pt}, toprule={0pt}, bottomrule={0pt}, left={10pt}, right={10pt}, top={5pt}, bottom={5pt},breakable]
1. \textbf{Disease Name:} \\
   - Clearly state the full name of the disease, including common aliases. \\
2. \textbf{Disease Description:} \\
   - Provide a concise and detailed definition or background information, covering causes, symptoms, and related important features. \\
3. \textbf{Disease Classification:} \\
   - Mention the medical classification of the disease, such as internal medicine, respiratory medicine, pediatrics, etc. \\
4. \textbf{Preventive Measures:} \\
   - List feasible and effective preventive methods. \\
5. \textbf{Causes of Illness:} \\
   - Clarify the main pathogenic factors and related triggers. \\
6. \textbf{Pathogenesis:} \\
   - Briefly explain the biological mechanism or pathological process of the disease. \\
7. \textbf{Symptoms:} \\
   - List specific and common symptoms or signs. \\
8. \textbf{Insurance Status:} \\
   - Indicate whether the disease is covered by health insurance. \\
9. \textbf{Incidence Probability:} \\
   - Provide data on the incidence or probability of occurrence of the disease. \\
10. \textbf{Susceptible Population:} \\
    - Identify high-risk groups or those prone to the disease. \\
11. \textbf{Transmission Method:} \\
    - Explain the main transmission routes of the disease; if non-contagious, specify as such. \\
12. \textbf{Comorbid Conditions:} \\
    - Mention other health problems or complications that may coexist with this disease. \\
13. \textbf{Consultation Department:} \\
    - Recommend the medical department patients should visit. \\
14. \textbf{Treatment Methods:} \\
    - Describe the main treatment approaches, including both pharmacological and non-pharmacological treatments. \\
15. \textbf{Treatment Duration:} \\
    - Provide an approximate time range for treatment. \\
16. \textbf{Cure Probability:} \\
    - Provide data on the cure rate for the disease. \\
17. \textbf{Common Medications:} \\
    - List common medications used to treat this disease. \\
18. \textbf{Treatment Costs:} \\
    - Provide an estimated cost range for treating this disease. \\
19. \textbf{Diagnostic Tests:} \\
    - List possible examinations and tests needed for diagnosing this disease. \\
20. \textbf{Recommended Diet:} \\
    - Provide a list of foods suggested for patients. \\
21. \textbf{Prohibited Foods:} \\
    - Provide a list of foods patients should avoid. \\
22. \textbf{Recommended Recipes:} \\
    - Provide one or two suitable recipes or food combinations for patients. \\
23. \textbf{Recommended Medications:} \\
    - List specific brands or manufacturers of recommended medications. \\
24. \textbf{Detailed Medication Information:} \\
    - Provide detailed information on some recommended medications, including drug names and manufacturers.
\end{tcolorbox}

\noindent\textbf{Refined Criterion for Question: ``What are the emergency treatment methods for allergic shock?''}
\begin{tcolorbox}[colback=gray!10, colframe=gray!100, rounded corners, leftrule={0pt}, rightrule={0pt}, toprule={0pt}, bottomrule={0pt}, left={10pt}, right={10pt}, top={5pt}, bottom={5pt},breakable]
1. \textbf{Disease Name:} \\
   - Clearly state the full name of the disease: Anaphylactic Shock \\
2. \textbf{Disease Description:} \\
   - Provide a concise and detailed definition or background information: Anaphylactic shock is a systemic rapid allergic reaction primarily characterized by acute peripheral circulation insufficiency, triggered by specific allergens that are generally harmless to the human body in patients with an allergic constitution. \\
   - Cover causes, symptoms, and related important features: Causes include antibiotics, local anesthetics, vitamins, etc.; symptoms include skin itching, hives, difficulty breathing, chest tightness, coughing, abdominal pain, nausea, vomiting, dizziness, and pallor. \\
3. \textbf{Disease Classification:} \\
   - Mention the medical classification of the disease: Medical Encyclopedia, Emergency Medicine \\
4. \textbf{Preventive Measures:} \\
   - List feasible and effective preventive methods: Consult detailed allergy history before medication, minimize unnecessary injections, prefer oral formulations, and observe allergic patients for 15–20 minutes after injections. Conduct skin prick tests to avoid using medications that may yield positive results. \\
5. \textbf{Causes of Illness:} \\
   - Clarify the main pathogenic factors and related triggers: Endocrine substances, enzymes, pollen extracts, food, antisera, occupational exposure proteins, insect venoms, etc. \\
6. \textbf{Pathogenesis:} \\
   - Briefly explain the biological mechanism or pathological process: External antigenic substances entering the body can stimulate the immune system to produce corresponding antibodies. The production of IgE varies significantly among individuals due to genetic differences. These specific IgE have a strong affinity for target cells in the skin, bronchi, blood vessel walls, etc. When the same antigen contacts a sensitized individual again, it can trigger a widespread Type I hypersensitivity reaction. \\
7. \textbf{Symptoms:} \\
   - List specific and common symptoms or signs: Pale lips, fainting, altered consciousness, hives, consciousness disorders, palpitations, hypotension, etc. \\
8. \textbf{Insurance Status:} \\
   - Indicate whether the disease is covered by health insurance: No \\
9. \textbf{Incidence Probability:} \\
   - Provide data on the incidence or probability of occurrence of the disease: 0.001\% \\
10. \textbf{Susceptible Population:} \\
    - Identify high-risk groups or those prone to the disease: Individuals with an allergic constitution \\
11. \textbf{Transmission Method:} \\
    - Explain the main transmission routes of the disease; if non-contagious, specify as such: Non-contagious \\
12. \textbf{Comorbid Conditions:} \\
    - Mention other health problems or complications that may coexist with this disease: Edema \\
13. \textbf{Consultation Department:} \\
    - Recommend the medical department patients should visit: Emergency Department \\
14. \textbf{Treatment Methods:} \\
    - Describe the main treatment approach: Immediate on-site rescue upon occurrence \\
15. \textbf{Treatment Duration:} \\
    - Provide an approximate time range for treatment: Depends on the specific situation \\
16. \textbf{Cure Probability:} \\
    - Provide data on the cure rate for the disease: 80\% \\
17. \textbf{Common Medications:} \\
    - List common medications used to treat this disease: Diphenhydramine hydrochloride injection, Naloxone hydrochloride injection, Epinephrine hydrochloride injection, Dobutamine glucose injection, Hydrocortisone injection, Dextrose 40 injection, Sodium hydrocortisone succinate for injection \\
18. \textbf{Treatment Costs:} \\
    - Provide an estimated cost range for treating this disease: Approximately (5000-10000 yuan) at a city-level top-tier hospital \\
19. \textbf{Diagnostic Tests:} \\
    - List possible examinations and tests needed for diagnosing this disease: Electrocardiogram, Free heparin time, Complete blood count, Blood biochemical panel, Renal function tests, Intradermal enzyme tests, Heparin-antithrombin IV complex, Streptomycin intradermal test, Central venous pressure measurement (CVP) \\
20. \textbf{Recommended Diet:} \\
    - Provide a list of foods suggested for patients: Eggs, quail eggs, sesame seeds, pumpkin seeds \\
21. \textbf{Prohibited Foods:} \\
    - Provide a list of foods patients should avoid: Honey, white lentils, crabs, lard (rendered fat) \\
22. \textbf{Recommended Recipes:} \\
    - Provide one or two suitable recipes or food combinations for patients: None \\
23. \textbf{Recommended Medications:} \\
    - List specific brands or manufacturers of recommended medications: Suicheng Pharmaceutical Diphenhydramine hydrochloride injection, Fuan Pharmaceutical Group Sodium hydrocortisone succinate for injection, Shiyao Silver Lake Epinephrine hydrochloride injection, Wuzhong Pharmaceutical Naloxone hydrochloride injection, Jinyao Pharmaceutical Epinephrine hydrochloride injection, Beijing Yongkang Pharmaceutical Epinephrine hydrochloride injection, Sunuo Naloxone hydrochloride injection, Haiwangfuyuan Dobutamine glucose injection, Aierhaitai Naloxone hydrochloride injection, Disainuo Pharmaceutical Naloxone hydrochloride injection, Changzheng Fumin Jingshan Dextrose 40 injection, Jiangsu Yuanheng Dextrose 40 injection, Yunchong Pharmaceutical Dextrose 40 injection, Tianfeng Pharmaceutical Dextrose 40 injection, Yuandong Pharmaceutical Naloxone hydrochloride injection, Beijing Yongkang Pharmaceutical Diphenhydramine hydrochloride injection, Lijun Pharmaceutical Hydrocortisone injection, Tianyi Qinkun Dextrose 40 injection, Baiyunshan Mingxing Epinephrine hydrochloride injection, Yue Pharmaceutical Dextrose 40 injection, Dikan Changjiang Pharmaceutical Epinephrine hydrochloride injection, Chengdu Tiantai Mountain Pharmaceutical Naloxone hydrochloride injection, Yonghe Pharmaceutical Dextrose 40 injection \\
24. \textbf{Detailed Medication Information:} \\
    - Provide detailed information on some recommended medications, including drug names and manufacturers: Suicheng Pharmaceutical Diphenhydramine hydrochloride injection, Fuan Pharmaceutical Group Sodium hydrocortisone succinate for injection, Shiyao Silver Lake Epinephrine hydrochloride injection, Wuzhong Pharmaceutical Naloxone hydrochloride injection, Jinyao Pharmaceutical Epinephrine hydrochloride injection, Beijing Yongkang Pharmaceutical Epinephrine hydrochloride injection, Sunuo Naloxone hydrochloride injection, Haiwangfuyuan Dobutamine glucose injection, Aierhaitai Naloxone hydrochloride injection, Disainuo Pharmaceutical Naloxone hydrochloride injection, Changzheng Fumin Jingshan Dextrose 40 injection, Jiangsu Yuanheng Dextrose 40 injection, Yunchong Pharmaceutical Dextrose 40 injection, Tianfeng Pharmaceutical Dextrose 40 injection, Yuandong Pharmaceutical Naloxone hydrochloride injection, Beijing Yongkang Pharmaceutical Diphenhydramine hydrochloride injection, Lijun Pharmaceutical Hydrocortisone injection, Tianyi Qinkun Dextrose 40 injection, Baiyunshan Mingxing Epinephrine hydrochloride injection, Yue Pharmaceutical Dextrose 40 injection, Dikan Changjiang Pharmaceutical Epinephrine hydrochloride injection, Chengdu Tiantai Mountain Pharmaceutical Naloxone hydrochloride injection, Yonghe Pharmaceutical Dextrose 40 injection.
\end{tcolorbox}

Below, we present the 6-turn interaction starting from the initial question ``What are the emergency treatment methods for allergic shock?''
\begin{tcolorbox}[colback=gray!10, colframe=gray!100, rounded corners, leftrule={0pt}, rightrule={0pt}, toprule={0pt}, bottomrule={0pt}, left={12pt}, right={12pt}, top={8pt}, bottom={8pt},breakable]
---------------- Turn 0 ----------------\\
\textbf{Initial Question: }What are the emergency treatment methods for allergic shock? \\
\textbf{Full marks: } 1. Disease Name: 1 point; 2. Disease Description: 2 points; 3. Disease Classification: 1 point; 4. Preventive Measures: 1 point; 5. Causes of Disease: 1 point; 6. Pathogenesis: 1 point; 7. Symptoms: 1 point; 8. Insurance Status: 1 point; 9. Incidence Probability: 1 point; 10. Vulnerable Population: 1 point; 11. Transmission Method: 1 point; 12. Comorbidities: 1 point; 13. Consultation Department: 1 point; 14. Treatment Method: 1 point; 15. Treatment Duration: 1 point; 16. Cure Probability: 1 point; 17. Common Medications: 1 point; 18. Treatment Cost: 1 point; 19. Examination Items: 1 point; 20. Suitable Diet: 1 point; 21. Forbidden Diet: 1 point; 22. Recommended Recipes: 1 point; 23. Recommended Medications: 1 point; 24. Medication Details: 1 point.
\\
\textbf{Answer: }Allergic shock is a severe allergic reaction that usually requires emergency treatment. Here are the common emergency treatment methods for allergic shock: Stay calm, have the patient lie flat, loosen the collar and belt, and ensure the airway is clear. Immediately administer intravenous epinephrine to dilate blood vessels and increase blood pressure. Monitor vital signs such as blood pressure, heart rate, and respiration, and record the time for the doctor's diagnosis and treatment. If the patient has difficulty breathing, perform artificial respiration and initiate cardiopulmonary resuscitation. Transport the patient to the hospital for treatment, including fluid and blood transfusion support. It is important to note that if the patient experiences severe allergic reactions, such as difficulty breathing, rapid heartbeat, or confusion, call emergency services or take them to the hospital immediately.\\  
\textbf{Summarized Answer:} Allergic shock is a severe allergic reaction that requires emergency treatment. Common emergency treatment methods include: Stay calm, have the patient lie flat, loosen the collar and belt, and ensure the airway is clear. Immediately administer intravenous epinephrine to dilate blood vessels and increase blood pressure. Monitor the patient's blood pressure, heart rate, and respiration, and record the time for the doctor's diagnosis and treatment. If the patient has difficulty breathing, perform artificial respiration and initiate cardiopulmonary resuscitation. Transport the patient to the hospital for treatment, including fluid and blood transfusion support. It is important to note that if the patient experiences severe allergic reactions, such as difficulty breathing, rapid heartbeat, or confusion, call emergency services or take them to the hospital immediately.
\\
\textbf{Score: } 1. Disease Name: 1/1; 2. Disease Description: 0/2; 3. Disease Classification: 0/1; 4. Preventive Measures: 0/1; 5. Causes of Disease: 0/1; 6. Pathogenesis: 0/1; 7. Symptoms: 0/1; 8. Insurance Status: 0/1; 9. Incidence Probability: 0/1; 10. Vulnerable Population: 0/1; 11. Transmission Method: 0/1; 12. Comorbidities: 0/1; 13. Consultation Department: 0/1; 14. Treatment Method: 1/1; 15. Treatment Duration: 0/1; 16. Cure Probability: 0/1; 17. Common Medications: 0/1; 18. Treatment Cost: 0/1; 19. Examination Items: 0/1; 20. Suitable Diet: 0/1; 21. Forbidden Diet: 0/1; 22. Recommended Recipes: 0/1; 23. Recommended Medications: 0/1; 24. Medication Details: 0/1. \\
\textbf{Total Score: } 2/25.\\
---------------- Turn 1 ----------------\\
\textbf{Follow-up Question: }What are the causes of allergic shock?\\
\textbf{Answer: }The causes of allergic shock may be due to allergic reactions, which can include: drug allergies (certain medications such as antibiotics, hormones, anticoagulants, non-steroidal anti-inflammatory drugs, etc.), food allergies (such as eggs, peanuts, milk, etc.), environmental factors (such as climate changes, air pressure changes, pollen, etc.), and psychological factors (such as emotional fluctuations, anxiety, stress, etc.). In summary, allergic shock is a serious allergic reaction that requires emergency treatment.\\
\textbf{Summarized Answer: }Allergic shock is a severe allergic reaction that usually requires emergency treatment. Common emergency treatment methods include: remain calm, have the patient lie flat, loosen the collar and belt to ensure the airway is clear; immediately administer intravenous epinephrine to dilate blood vessels and increase blood pressure; monitor vital signs such as blood pressure, heart rate, and respiration, and record the time for the doctor's diagnosis and treatment. If the patient has difficulty breathing, perform artificial respiration and initiate cardiopulmonary resuscitation. Transport the patient to the hospital for treatment, including fluid and blood transfusion support. It is important to note that if the patient experiences severe allergic reactions, such as difficulty breathing, rapid heartbeat, or confusion, call emergency services or take them to the hospital immediately.\\
\textbf{Score: } 1. Disease Name: 1/1; 2. Disease Description: 0/2; 3. Disease Classification: 0/1; 4. Preventive Measures: 0/1; 5. Causes of Disease: 1/1; 6. Pathogenesis: 0/1; 7. Symptoms: 0/1; 8. Insurance Status: 0/1; 9. Incidence Probability: 0/1; 10. Vulnerable Population: 0/1; 11. Transmission Method: 0/1; 12. Comorbidities: 0/1; 13. Consultation Department: 0/1; 14. Treatment Method: 1/1; 15. Treatment Duration: 0/1; 16. Cure Probability: 0/1; 17. Common Medications: 0/1; 18. Treatment Cost: 0/1; 19. Examination Items: 0/1; 20. Suitable Diet: 0/1; 21. Forbidden Diet: 0/1; 22. Recommended Recipes: 0/1; 23. Recommended Medications: 0/1; 24. Medication Details: 0/1.\\
\textbf{Total Score: }3/25.\\
---------------- Turn 2 ----------------\\
\textbf{Follow-up Question: }What is the pathogenesis of allergic shock?\\
\textbf{Answer: }The pathogenesis of allergic shock is quite complex and involves multiple factors. Some common mechanisms include: 1. Allergic reactions: Allergens are a significant cause of allergic shock, which may include certain drugs, foods, and pollen. 2. Immune system abnormalities: Allergic reactions can be caused by immune system dysfunction, such as allergic asthma and allergic rhinitis. 3. Vascular spasms: Allergic reactions can also lead to vascular spasms, contributing to conditions like allergic shock and allergic dermatitis. 4. Neural reflexes: Allergic reactions may involve responses from the nervous system, as seen in allergic shock and allergic asthma. In summary, the pathogenesis of allergic shock is complex and requires comprehensive analysis and diagnosis based on specific circumstances.\\
\textbf{Summarized Answer: }Allergic shock is a severe allergic reaction that usually requires emergency treatment. Common emergency treatment methods include: remain calm, have the patient lie flat, loosen the collar and belt to ensure the airway is clear; immediately administer intravenous epinephrine to dilate blood vessels and increase blood pressure; monitor vital signs such as blood pressure, heart rate, and respiration, and record the time for the doctor's diagnosis and treatment. If the patient has difficulty breathing, perform artificial respiration and initiate cardiopulmonary resuscitation. Transport the patient to the hospital for treatment, including fluid and blood transfusion support. It is important to note that if the patient experiences severe allergic reactions, such as difficulty breathing, rapid heartbeat, or confusion, call emergency services or take them to the hospital immediately.\\
\textbf{Score:} 1. Disease Name: 1/1; 2. Disease Description: 0/2; 3. Disease Classification: 0/1; 4. Preventive Measures: 0/1; 5. Causes of Disease: 1/1; 6. Pathogenesis: 0.5/1; 7. Symptoms: 0/1; 8. Insurance Status: 0/1; 9. Incidence Probability: 0/1; 10. Vulnerable Population: 0/1; 11. Transmission Method: 0/1; 12. Comorbidities: 0/1; 13. Consultation Department: 0/1; 14. Treatment Method: 1/1; 15. Treatment Duration: 0/1; 16. Cure Probability: 0/1; 17. Common Medications: 0/1; 18. Treatment Cost: 0/1; 19. Examination Items: 0/1; 20. Suitable Diet: 0/1; 21. Forbidden Diet: 0/1; 22. Recommended Recipes: 0/1; 23. Recommended Medications: 0/1; 24. Medication Details: 0/1.\\
\textbf{Total Score: }3.5/25.\\
---------------- Turn 3 ----------------\\
\textbf{Challenge: }Are vascular spasms and neural reflexes the main mechanisms of pathogenesis in allergic shock?\\
\textbf{Answer: }Yes, the pathogenesis of allergic shock is quite complex and involves multiple factors. Specifically, vascular spasms and neural reflexes are among the main mechanisms of allergic shock. The mechanisms include: 1. Allergic reactions: Allergens are a significant cause, which may include certain drugs, foods, and pollen. 2. Immune system abnormalities: Allergic reactions can result from immune system dysfunction, such as allergic asthma and allergic rhinitis. 3. Vascular spasms: Allergic reactions may lead to vascular spasms, contributing to conditions like allergic shock and allergic dermatitis. 4. Neural reflexes: Responses from the nervous system may also play a role, as seen in allergic shock and allergic asthma. In summary, the pathogenesis of allergic shock is complex and requires comprehensive analysis and diagnosis based on specific circumstances.\\
\textbf{Summarized Answer: }Allergic shock is a severe allergic reaction that usually requires emergency treatment. Common emergency treatment methods include: remain calm, have the patient lie flat, loosen the collar and belt to ensure the airway is clear; immediately administer intravenous epinephrine to dilate blood vessels and increase blood pressure; monitor vital signs such as blood pressure, heart rate, and respiration, and record the time for the doctor's diagnosis and treatment. If the patient has difficulty breathing, perform artificial respiration and initiate cardiopulmonary resuscitation. Transport the patient to the hospital for treatment, including fluid and blood transfusion support. It is important to note that if the patient experiences severe allergic reactions, such as difficulty breathing, rapid heartbeat, or confusion, call emergency services or take them to the hospital immediately.\\
\textbf{Evaluation:} 1. Disease Name: 1/1; 2. Disease Description: 0/2; 3. Disease Classification: 0/1; 4. Preventive Measures: 0/1; 5. Causes of Disease: 1/1; 6. Pathogenesis: 0.5/1; 7. Symptoms: 0/1; 8. Insurance Status: 0/1; 9. Incidence Probability: 0/1; 10. Vulnerable Population: 0/1; 11. Transmission Method: 0/1; 12. Comorbidities: 0/1; 13. Consultation Department: 0/1; 14. Treatment Method: 1/1; 15. Treatment Duration: 0/1; 16. Cure Probability: 0/1; 17. Common Medications: 0/1; 18. Treatment Cost: 0/1; 19. Examination Items: 0/1; 20. Suitable Diet: 0/1; 21. Forbidden Diet: 0/1; 22. Recommended Recipes: 0/1; 23. Recommended Medications: 0/1; 24. Medication Details: 0/1.\\
\textbf{Total Score: }3.5/25.\\
---------------- Turn 4 ----------------\\
\textbf{Follow-up Question: }What are the symptoms of allergic shock?\\
\textbf{Answer: }The main symptoms of allergic shock include: difficulty breathing, rapid heartbeat, confusion, facial and limb cyanosis, nausea, vomiting, diarrhea, dizziness, flushing, sweating, abnormal heart function (increased or decreased blood pressure), respiratory symptoms (like allergic asthma and rhinitis), and neurological symptoms (such as confusion, seizures, and coma). It is important to note that the causes of allergic shock can be diverse and require comprehensive analysis and diagnosis.\\
\textbf{Summarized Answer: }Allergic shock is a severe allergic reaction that requires emergency treatment. Common emergency treatment methods include: keeping the patient calm, having them lie flat, loosening their collar and belt to maintain airway clearance; immediately administering intravenous epinephrine to dilate blood vessels and raise blood pressure; monitoring vital signs such as blood pressure, heart rate, and respiration while recording the time for proper medical assessment. If the patient experiences difficulty breathing, perform artificial respiration and initiate cardiopulmonary resuscitation. The patient should be transported to a hospital for supportive treatment, including fluid and blood transfusions. If severe symptoms like difficulty breathing, rapid heartbeat, or confusion occur, call emergency services or take the patient to the hospital immediately.\\
\textbf{Evaluation:} 1. Disease Name: 1/1; 2. Disease Description: 1.5/2; 3. Disease Classification: 0/1; 4. Preventive Measures: 0/1; 5. Causes of Disease: 1/1; 6. Pathogenesis: 0.5/1; 7. Symptoms: 1/1; 8. Insurance Status: 0/1; 9. Incidence Probability: 0/1; 10. Vulnerable Population: 0/1; 11. Transmission Method: 0/1; 12. Comorbidities: 0/1; 13. Consultation Department: 0/1; 14. Treatment Method: 1/1; 15. Treatment Duration: 0/1; 16. Cure Probability: 0/1; 17. Common Medications: 0/1; 18. Treatment Cost: 0/1; 19. Examination Items: 0/1; 20. Suitable Diet: 0/1; 21. Forbidden Diet: 0/1; 22. Recommended Recipes: 0/1; 23. Recommended Medications: 0/1; 24. Medication Details: 0/1.\\
\textbf{Total Score: }6/25.\\
---------------- Turn 5 ----------------\\
\textbf{Follow-up Question: }What are the specific physiological processes involved in the pathogenesis of allergic shock besides allergic reactions and immune system abnormalities?\\
\textbf{Answer: }The pathogenesis of allergic shock is complex and requires comprehensive analysis and diagnosis. Common mechanisms include: 1. Allergic reactions triggered by allergens such as drugs, food, or pollen; 2. Abnormal immune responses; 3. Vascular spasms, which may lead to conditions like allergic shock and dermatitis; 4. Neural reflexes related to the nervous system; 5. Neurological symptoms that can arise from the nervous system's influence; and 6. Respiratory symptoms are associated with allergic reactions. Each case needs individual assessment.\\
\textbf{Summarized Answer: }Allergic shock is a serious allergic reaction that usually necessitates emergency treatment. Common emergency interventions include keeping the patient calm, having them lie flat, loosening clothing to ensure airway clearance, and immediately administering intravenous epinephrine to dilate blood vessels and elevate blood pressure. Vital signs should be monitored and recorded for further medical evaluation. If breathing difficulties arise, perform artificial respiration and initiate cardiopulmonary resuscitation. Transport the patient to the hospital for additional treatments like fluid and blood transfusions. It is crucial to act immediately if severe symptoms like difficulty breathing, rapid heartbeat, or confusion occur.\\
\textbf{Evaluation:} 1. Disease Name: 1/1; 2. Disease Description: 1.5/2; 3. Disease Classification: 0/1; 4. Preventive Measures: 0/1; 5. Causes of Disease: 1/1; 6. Pathogenesis: 0.5/1; 7. Symptoms: 1/1; 8. Insurance Status: 0/1; 9. Incidence Probability: 0/1; 10. Vulnerable Population: 0/1; 11. Transmission Method: 0/1; 12. Comorbidities: 0/1; 13. Consultation Department: 0/1; 14. Treatment Method: 1/1; 15. Treatment Duration: 0/1; 16. Cure Probability: 0/1; 17. Common Medications: 0/1; 18. Treatment Cost: 0/1; 19. Examination Items: 0/1; 20. Suitable Diet: 0/1; 21. Forbidden Diet: 0/1; 22. Recommended Recipes: 0/1; 23. Recommended Medications: 0/1; 24. Medication Details: 0/1.\\
\textbf{Total Score: }6/25.
\end{tcolorbox}

\section{Prompt}
\label{appendix:prompt}
\noindent \textbf{Prompt for General Domain-level Criteria}
\begin{tcolorbox}[colback=gray!10, colframe=gray!100, rounded corners, leftrule={0pt}, rightrule={0pt}, toprule={0pt}, bottomrule={0pt}, left={10pt}, right={10pt}, top={5pt}, bottom={5pt},breakable]
You are an expert in \{Domain\}. Below are some questions, reference answers, and apprentice responses for the \{Topic\} scenario: \{Examples\}.

Please summarize the shortcomings of the apprentice's response in one sentence. Finally, provide a brief list of criteria that others can use to evaluate responses to similar questions.
\end{tcolorbox}

\noindent \textbf{Prompt for Refined Question-level Criteria}
\begin{tcolorbox}[colback=gray!10, colframe=gray!100, rounded corners, leftrule={0pt}, rightrule={0pt}, toprule={0pt}, bottomrule={0pt}, left={10pt}, right={10pt}, top={5pt}, bottom={5pt},breakable]
You are an expert in \{Domain\}. Below is the knowledge base for the question \{Question\}.

Knowledge Base:
\{Retrieved Context\}.

Your task is to refine the following response criteria based on the knowledge base, ignoring any aspects not covered in it.

Criteria:
\{General Criteria\}.

Please refine the criteria related to answering the question based on the contents of the knowledge base, clearly listing the specific elements that the response should include in order to evaluate its performance. Do not simplify the content; only list the refined criteria without explanations or responses.
\end{tcolorbox}

\noindent \textbf{Prompt for Scoring}
\begin{tcolorbox}[colback=gray!10, colframe=gray!100, rounded corners, leftrule={0pt}, rightrule={0pt}, toprule={0pt}, bottomrule={0pt}, left={10pt}, right={10pt}, top={5pt}, bottom={5pt},breakable]
You are an expert in \{Domain\}. Below is the response from the model to be evaluated. Your task is to assess the model's capability. The evaluation criteria are as follows:  
\{Refined Criteria\}.

Model Response:  
\{Answer\}.

Please first list the criteria and the maximum score for each, with the maximum score being the number of key pieces of information for that criterion. Then, evaluate and score the response based on each criterion, providing the score/max score for each. Scoring can be in decimal form. Finally, present the total score/max score as a numerical value. Do not include greetings.  
Your output format should be: \\
\{``Criteria and Full Marks'':  ``'', ``Evaluation'':  ``''\}.
\end{tcolorbox}

\noindent \textbf{Prompt for Initial Questions}
\begin{tcolorbox}[colback=gray!10, colframe=gray!100, rounded corners, leftrule={0pt}, rightrule={0pt}, toprule={0pt}, bottomrule={0pt}, left={10pt}, right={10pt}, top={5pt}, bottom={5pt},breakable]
You are a user of the \{Domain\} model. Please generate different user questions for the \{Topic\} scenario based on the context to evaluate the model's understanding of \{Domain\}. Ensure that the questions are fact-based, can be answered from the context, and are logically structured. List the questions directly without providing answers.

Relevant context:
\{Retrieved Context\}.
\\
Useful User Questions:
\end{tcolorbox}

\noindent \textbf{Prompt for Follow-up Questions}
\begin{tcolorbox}[colback=gray!10, colframe=gray!100, rounded corners, leftrule={0pt}, rightrule={0pt}, toprule={0pt}, bottomrule={0pt}, left={10pt}, right={10pt}, top={5pt}, bottom={5pt},breakable]
Please simulate a user of the \{Domain\} model. Based on the dialogue history and the scores assigned to the model's responses according to the criteria, generate a follow-up question targeting the areas with low scores in order to assess the model's understanding of \{Domain\}. The follow-up question should be answerable based on the criteria. Your follow-up should relate to the dialogue history.

Criteria:  
\{Refined Criteria\}.

Dialogue:  
\{History\}.

Score:  
\{Score\}.

Instructions:  
Your follow-up should aim to reference the model's responses while maintaining coherence in the dialogue and enhancing its interconnectedness. Please note that it is important not to disclose the content of the criteria. Do not include greetings; simply provide the follow-up question. Your question should be answerable based on the criteria.
\end{tcolorbox}

\noindent \textbf{Prompt for Challenges}
\begin{tcolorbox}[colback=gray!10, colframe=gray!100, rounded corners, leftrule={0pt}, rightrule={0pt}, toprule={0pt}, bottomrule={0pt}, left={10pt}, right={10pt}, top={5pt}, bottom={5pt},breakable]
You are a user of the \{Domain\} model. Below is the dialogue between you and the model, along with the evaluation criteria. Please pose a challenging question to the model in the form of ``Are you sure your answer is correct? ''

Criteria:  
\{Refined Criteria\}.

Dialogue:
\{History\}.

Instructions:
Your challenge should relate to the previous question and the criteria. It should be based on your historical questions while maintaining coherence in the dialogue and enhancing its interconnectedness. Do not include greetings; simply provide your challenge. Ensure that your challenge can be answered based on the evaluation criteria.
\end{tcolorbox}

\noindent \textbf{Prompt for Qualitative Evaluation}
\begin{tcolorbox}[colback=gray!10, colframe=gray!100, rounded corners, leftrule={0pt}, rightrule={0pt}, toprule={0pt}, bottomrule={0pt}, left={10pt}, right={10pt}, top={5pt}, bottom={5pt},breakable]
You are a \{domain\} expert, and your task is to provide a comprehensive evaluation of the model's capabilities. You need to assess the model's responses in terms of professionalism and stability:\\

Professionalism Evaluation:\\
1. Factual Accuracy: Is the information provided by the model accurate and consistent with established standards? You may refer to the score given.\\
2. Meeting User Needs: Does the response meet the purpose and needs expressed by the user? Does it provide a complete and appropriate answer to the question?\\
3. Clarity: Is the response clear and easy to understand? Does it use simple language and structure to allow the user to easily grasp the information?\\

Stability Evaluation:\\
1. Consistency: Does the model maintain consistency in the information it provides across multiple responses?\\
2. Handling of Challenges: Does the model maintain the correctness of its answers when faced with a challenge?\\
3. Handling of Follow-up Questions: Is the model able to provide further information when prompted with follow-up questions and address the evaluator's inquiries?\\
Note that longer responses are not always better. Short and concise responses that meet the above requirements are preferred.\\

Evaluation Procedure: When you begin your evaluation, you need to follow the procedure below:

1. Evaluate the AI assistant's response based on the different dimensions. After evaluating each dimension, assign a score between 1 and 10 for that dimension.\\
2. Finally, based on your evaluations of each dimension, provide an overall score for the model's professionalism and stability, respectively, on a 1-10 scale.\\
3. Your scoring should be as strict as possible and adhere to the following scoring rules: Overall, the higher the quality of the model's response, the higher the score.\\
-If the model's response contains irrelevant information, fundamental factual errors, or harmful content, the total score must be 1 to 2 points.\\
-If the model's response contains no severe errors and is mostly harmless but of poor quality and does not meet user needs, the total score should be 3 to 4 points.\\
-If the model's response generally meets the user's needs but performs poorly in some dimensions and is of medium quality, the total score should be 5 to 6 points.\\
-If the model's response is close in quality to the reference answer and performs well in all dimensions, the total score should be 7 to 8 points.\\
-Only if the model's response significantly exceeds the reference answer in quality, fully addressing the user's problem and all needs, and performing close to perfect across all dimensions, should the total score be 9 to 10 points.\\
Remember: You must provide an evaluation and explanation before assigning a score. After explaining each dimension, include the score for that dimension. Then, at the end of your response, return all your scores in the following dictionary format (including brackets), ensuring that all your scores are integers: \{\{`Dimension 1': score, `Dimension 2': score, ..., `Overall Score': score\}\}. For example: \{\{`Consistency': 9, `Handling of Challenges': 6, ..., `Stability Overall Score': 7\}\}, \{\{`Factual Accuracy': 9, `Meeting User Needs': 6, ..., `Professionalism Overall Score': 7\}\}.\\

Below is the conversation history and scoring, where the evaluator chooses to ask follow-up questions or challenge:\\
Conversation History and Scores: \{History and Scores\} \\
Evaluation Criteria: \{Criteria\}
\end{tcolorbox}

\end{document}